\pdfoutput = 1

\documentclass[11pt]{article}

\usepackage[]{EMNLP2023}

\usepackage{times}
\usepackage{latexsym}
\usepackage{amsfonts}
\usepackage{graphicx}
\usepackage{stmaryrd} 
\usepackage{multirow}
\usepackage{array}
\usepackage{booktabs}
\usepackage{makecell}
\newcommand{\mycomment}[1]{}
\usepackage{enumitem}
\usepackage{comment}
\usepackage{xcolor}
\usepackage{booktabs}
\usepackage{caption}
\usepackage{graphicx}

\usepackage{makecell}

\usepackage[T1]{fontenc}

\usepackage[utf8]{inputenc}

\usepackage{microtype}

\usepackage{inconsolata}
\usepackage{amsmath}

\title{RETROcode: Leveraging a Code Database for Improved Natural Language to Code Generation}

\author{Nathanaël Beau \and Benoît Crabbé \\
       Université de Paris, LLF, CNRS, 75013 Paris, France \\ 
       \texttt{nathanael.beau.gs@gmail.com} \\
       \texttt{benoit.crabbe@u-paris.fr} }
\begin{document}
\maketitle
\begin{abstract}
As text and code resources have expanded, large-scale pre-trained models have shown promising capabilities in code generation tasks, typically employing supervised fine-tuning with problem statement-program pairs. However, increasing model size and data volume for performance gains also raises computational demands and risks of overfitting. Addressing these challenges, we present RETROcode, a novel adaptation of the RETRO architecture \cite{RETRO} for sequence-to-sequence models, utilizing a large code database as an auxiliary scaling method. This approach, diverging from simply enlarging model and dataset sizes, allows RETROcode to leverage a vast code database for prediction, enhancing the model's efficiency by integrating extensive memory. Our findings indicate that RETROcode not only outperforms similar-sized traditional architectures on test sets but also approaches the effectiveness of the much larger Codex model, despite being trained from scratch on a substantially smaller dataset.
\end{abstract}

\section{Introduction}

Code generation is the task of automatically creating computer programs from natural language, generating potentially previously unseen code. It has a wide range of applications, from creating code snippets for developers to generating complete software applications. In recent years, the increasing availability of large amounts of code and natural language data has facilitated the development of powerful neural network models that can perform code generation with high accuracy.

One challenge in working with large amounts of natural language and code data is the lack of aligned examples, which require human expertise to annotate. To address this issue, one approach is to use large pre-trained models that have been trained on a large volume of code and/or natural language data, and then fine-tune them on the available annotated data \cite{pretrainingneubig, codet5, codex, alphacode}.
The use of large models with a high number of parameters can provide computational benefits during training and inference, as well as improved memorization of the training data. However, training these models can be computationally expensive, and the large number of parameters may lead to overfitting on the training data \cite{lmparrot, codexmemo}.

An alternative approach for translating natural language to code is code retrieval, which involves searching for and retrieving an appropriate code snippet from a code database \cite{coderetrieval, coderetrieval2, coderetrieval3}. However, these methods are becoming less commonly used as it is now possible to use pre-trained models that are trained on the entire code database and generate personalized code responses to a given query.

Methods for natural language generation often involve the use of generative models that are trained to associate text with data in a database. These solutions have two main advantages: they allow for the separation of world knowledge from language learning, and they enable the use of smaller model sizes. For instance, the Knn-Based Composite Memory system \cite{KNN-BasedCompositeMemory} assists a conversational agent by providing access to information from similar discussions and by supplying relevant knowledge from various sources based on the input user prompt. Another example is RETRO architecture \cite{RETRO}, which provides information to a language model as decoding goes using sentences similar to what was generated. In both cases, queries are made to a database by comparing the embeddings of the input or output with those in the database to obtain the nearest neighbours, and the resulting information is provided to the encoder or decoder, respectively.

In this paper, we introduce RETROcode, a transformer-based architecture that 
integrates a sequence-to-sequence architecture into \citet{RETRO}. This facilitates the simultaneous processing of dual inputs: natural language utterances and analogous code snippets retrieved from a database. Our strategy strives to harness the extensive available code data while minimizing the model parameters.

We present two methods for integrating this information within the decoder and conduct an in-depth analysis of the impact of various critical components on system performance. Our results outperform  architectures with an equivalent number of parameters and are the close to Codex's performance while trained from scratch on much less data and with significantly fewer parameters. This article delivers the following contributions:
\begin{itemize}
\item It establishes a novel transformer sequence-to-sequence architecture that combines information from natural language input and similar code from a database.
\item It investigates the impact of key architecture components on system performance, including database preprocessing, database code size, and two distinct methods for integrating database information into the decoder.
\item It proposes an effective hybrid database to not only take advantage of the large amounts of code available but also of natural language to code alignments, enabling LLM-like results on specialized tasks with little training data.
\end{itemize}

This article is organized as follows: In Section {\ref{sec:querytodatabase}}, we provide a formal description of the query system used to retrieve neighbours from the database during decoding. In Section \ref{model}, we detail our model architecture, including two methods for merging natural language intent with information retrieved from the database. In Section \ref{setup}, we highlight the critical elements of our architecture. Finally, in Section \ref{experiments}, we conduct experiments to examine the various key elements of our model, comparing 3 different approaches to generate code from natural language.

\section{Query Architecture}
\label{sec:querytodatabase}

\begin{figure}[htbp]
\begin{center}
\scalebox{0.5}{\includegraphics[width=\textwidth]{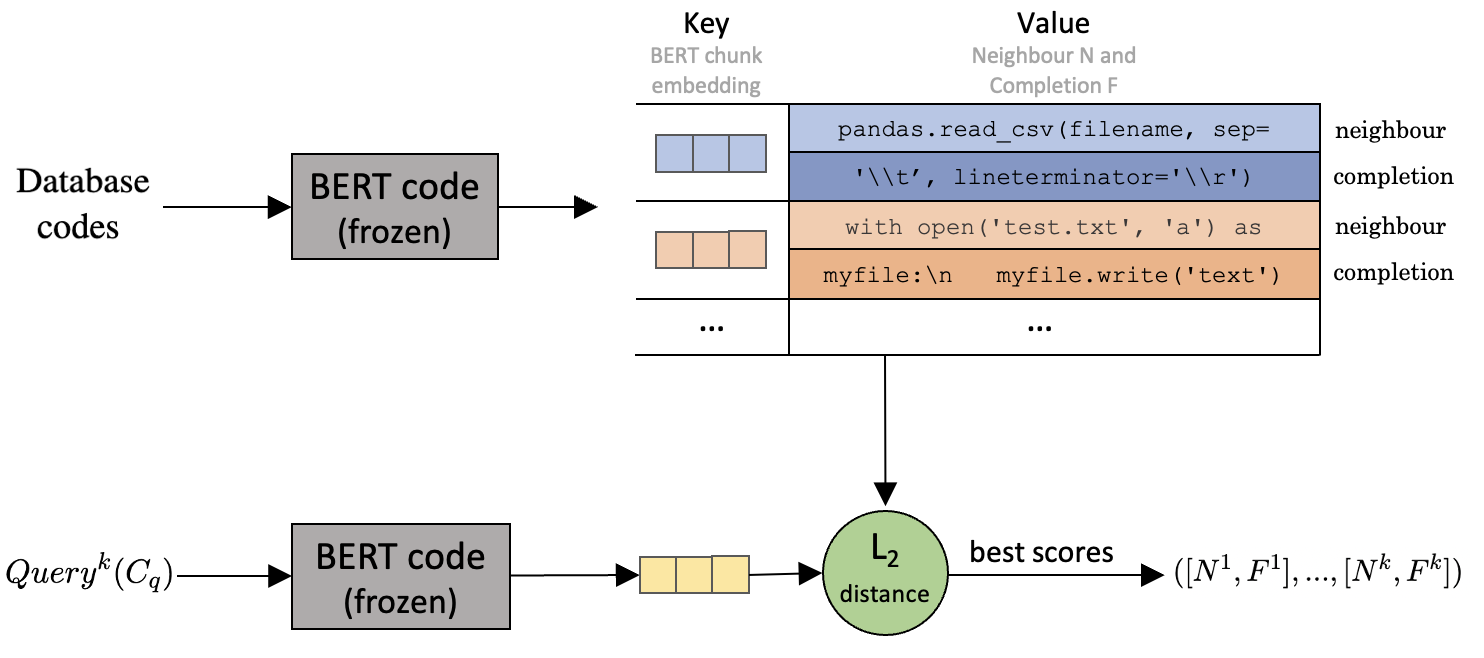}}
\caption{\label{fig:querytodatabase}Process of $Query_{k}(C_q)$ to obtain k-nearest neighbours and their continuation. Here, the chunk length $m$ to construct the database is equal to 8.}
\end{center}
\end{figure}

In this Section, we describe the database query system which is designed to retrieve similar codes from the database in response to a query as illustrated in Figure \ref{fig:querytodatabase}. The function $Query^{k}(C_q)$ is defined to take a code chunk of size $m$ and return its k-nearest code neighbours from the database. The embedding of the current code $C_q$ is calculated and compared to the embeddings from the database with an $L_2$ distance.  

We  first introduce the database organization and then explain how the query system is designed.

\subsection{Database structure}
\label{classicdatabase}

We structure our database $\mathcal{D}$ as a key-value memory. Each value consists of two continuous chunks of code tokens of size $m$, referred to as $[N, F]$. $N$ is the neighbouring chunk that is used to compute the key, while $F$ is the continuation of the code from $N$, adding  information. The key embedding is then computed with a frozen $\textsc{codeBERT}$ on $N$.

We choose to use a frozen model for the embedding calculation to optimize the efficiency of the database query system as it avoids  to re-compute embeddings over the entire database during training. It further enables the addition of new code chunks to the database after training.

Note that the concatenation $[N, F]$ is not necessarily a complete snippet of code, it depends of the size of $m$ which is one of the crucial parameters of our model. 
\\

\subsection{Neighbours Retrieval}

Given such a database, the query embedding of $C_q$  is also built with a frozen $\textsc{codeBERT}$. 
To retrieve the k-nearest neighbours and their continuations from ${\cal D}$, we use the $L_2$ distance:
\begin{equation*}
    Query_{k}(C_q)=(\mathcal{N}_1, ..., \mathcal{N}_k) \; \text{where} \; \mathcal{N}_i = [N_i, F_i]
\end{equation*}

Note that for a database of $T$ elements, we  query the approximate nearest neighbours in $O\log(T)$, relying on the Faiss library \citep{Faiss} \footnote{https://github.com/facebookresearch/faiss}.

\section{Model}
\label{model}

\subsection{Objective}

We consider a family of models that generate a code $Y$ from a natural language description $X$. The models have a generic form:
\begin{equation}
    p(Y \mid X) = \prod_{t} p(Y_t \mid Y_{<t}, X) 
\end{equation}
where $Y = \{Y_t : t \in \llbracket 1, L \rrbracket \}$ and $X = \{X_i : i \in \llbracket 1, n \rrbracket \}$.
The decoding objective aims to find the most-probable hypothesis among all candidate hypotheses by solving the following optimization problem:
\begin{equation}
\label{eqn:MAP}
\hat{Y} = \mathop{\text{argmax}}_{Y} p(Y \mid X)
\end{equation}

\subsection{Baseline}
\label{sec:baseline}

\begin{figure*}[htbp]
\begin{center}
\scalebox{1.}{
\includegraphics[width=\textwidth]{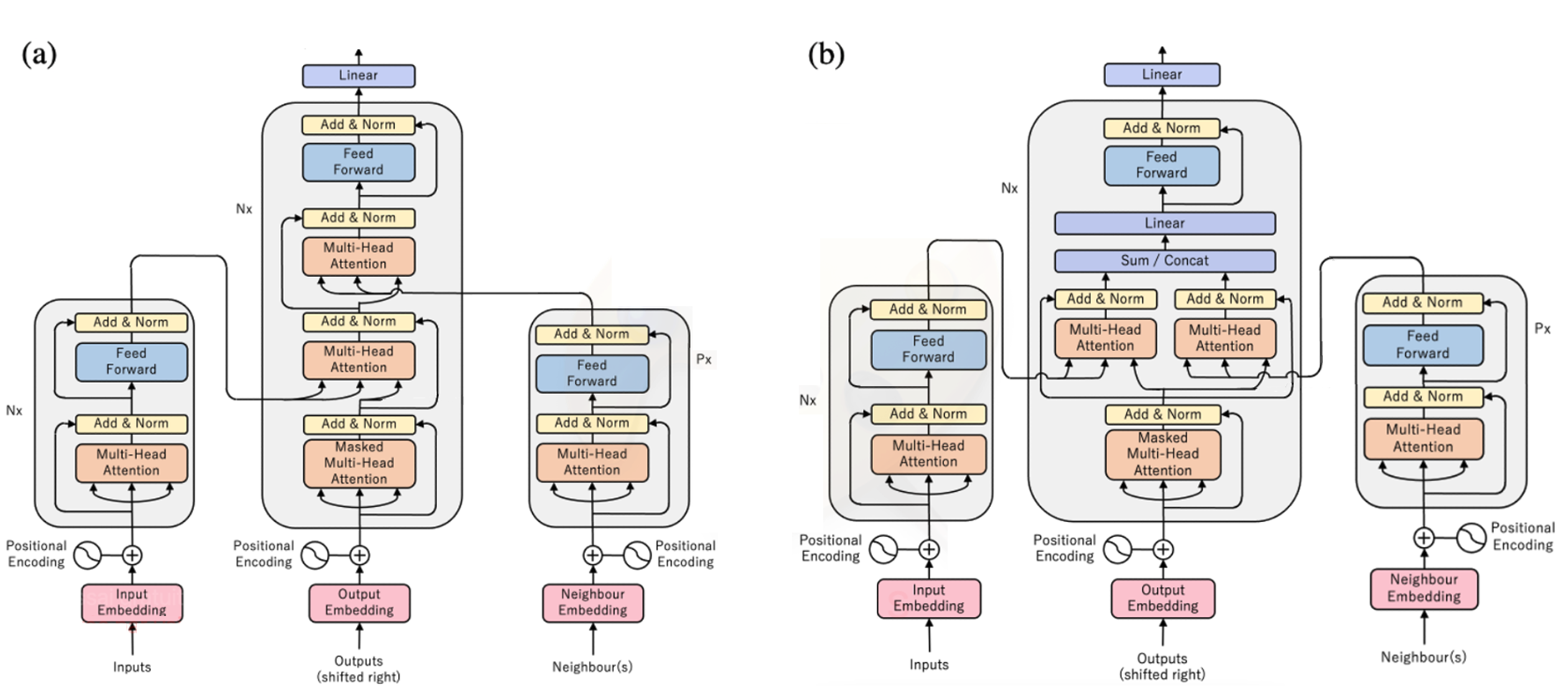}}
\end{center}
\caption{\label{fig:model}Illustration of the RETROcode architecture, which includes two variations for integrating neighbour encoding into the baseline model \footnotemark. (a) Sequential aggregation: we incorporate the information from the neighbours into the code generation process using a two-step process. First, we use the classic cross-attention mechanism to combine the information from the natural language. Then, we perform a second cross-attention between the output of the first cross-attention and the neighbours. This process is described in equation \ref{eqn:sequentialaggregation}.
(b) Parallel aggregation: we separately compute the information from the neighbours and the natural language with the decoder using cross-attention, and then merge the results with a linear layer as described in equation \ref{eqn:parallelaggregation}.}
\end{figure*}

We consider the classic transformer architecture from \citet{transformer} as our baseline with some minor changes: we replace Layer Normalisation with Root Mean Square (RMS) normalisation \cite{RMSnorm} and use rotary embedding \cite{rotaryemb}. As we employ residual connections \cite{residualconnection} between each sub-layer followed by a RMS normalization. We define  the notations:
\begin{equation*}
\label{eqn:utils}
\begin{split}
& \text{Sublayer}^{\textsc{ffw}}(X) = \text{RMSNorm}(X + \textsc{ffw}(X)) \\
& \text{Sublayer}^{\textsc{Sa}}(X) = \text{RMSNorm}(X + \textsc{Sa}(X)) \\
& \text{Sublayer}^{\textsc{Ca}}(X, Y) = \text{RMSNorm}(Y + \textsc{Ca}(X, Y)) \\
& \text{Sublayer}^{\textsc{Cca}}(X, Y) = \text{RMSNorm}(Y + \textsc{Cca}(X, Y))
\end{split}
\end{equation*}
where $\text{RMSNorm}$ is a Root Mean Square normalization, $\textsc{ffw}$ is a fully-connected feed-forward network. The self-attention \textsc{Sa} and the cross-attention \textsc{Ca} are classically defined as in \citet{transformer} with $\text{MultiHead}(Q, K, V)$ where $Q$, $K$, and $V$ are the query, key, and value matrices respectively. The chunked-cross attention {\sc Cca} is defined as in \citet{RETRO} to handle the interaction between the model and the retrieved data in chunks without breaking autoregressivity (see Appendix \ref{app:implementationdetails} for details). We can then define the encoder's layer for natural language as:
\begin{equation}
\label{eqn:baselineenc}
\begin{split}
\text{ENC\_NL}(X) & = \text{Sublayer}^{\textsc{FFW}}(H) \\
H &=\text{Sublayer}^{\textsc{Sa}}(X)
\end{split}
\end{equation}
and the code decoder's layer as follows:
\begin{equation}
\label{eqn:baselinedec}
\begin{split}
\text{DEC}(X, Y) & = \text{Sublayer}^{\textsc{FFW}}(C_{nl}) \\
C_{nl} &=\text{Sublayer}^{\textsc{Ca}}(E, C) \\
E &= \text{ENC}(X) \\
C &= \text{Sublayer}^{\textsc{Sa}}(Y)
\end{split}
\end{equation}
The methodology for computing hidden representations using a transformer has been detailed. To predict code tokens, our approach is to leverage a standard application of the softmax function across the model's vocabulary. However, to address the challenge of rare word terms, especially relevant in the context of very specific variable names, we enhance this with the inclusion of a pointer network \cite{pointernet}. In accordance with methodologies outlined in \citet{tranX, BERTranX}, the final output layer of our model is a fusion of a softmax distribution over the vocabulary and the results derived from the pointer network. This design ensures an effective balance between handling general language structures and accommodating specific OOV terms.

\subsection{Gathering neighbours' information}

Crucially, our proposed architecture incorporates a transformer guided by neighbours retrieved from an external code database. Here, we explain how we integrate information from the code database into the code generation process, that is the value returned by the $Query_k(C_q)$ function.

The information from the retrieved neighbours must be encoded to be integrated into the decoder. Each encoding of $\mathcal{N}_i$ is conditioned with the code already generated by the decoder ($Y$) as in \citet{RETRO}:
\begin{equation}
\label{eqn:neighbourencca}
\begin{split}
\text{ENC\_NB}(\mathcal{N}_i, Y) & = \text{Sublayer}^{\textsc{FFW}}(H) \\
H &=\text{Sublayer}^{\textsc{Ca}}(C, E) \\
C &= \text{Sublayer}^{\textsc{Sa}}(Y)\\
E &= \text{Sublayer}^{\textsc{Sa}}(\mathcal{N}_i) 
\end{split}
\end{equation}
As a result, the encoding $E_{nb}$ of the neighbours is the concatenation of the encoding of each retrieved neighbour: 
\begin{equation*}
\begin{split}
E_{nb} & = \text{ENC\_NB}(\mathcal{N}, Y)\\
&=[\text{ENC\_NB}(\mathcal{N}_1, Y):\ldots :\text{ENC\_NB}(\mathcal{N}_k, Y)]
\end{split}
\end{equation*}
Note that we cannot use directly the database key to feed our decoder since we wish to integrate not only similar codes but also their continuation, which are not included in the key computation.

\subsection{Decoding with natural language and neighbours} 
\label{RETROcodebase}

As we aim  to feed the decoder with information from natural language and the retrieved neighbours to guide upcoming predictions, we describe here an update of equation \ref{eqn:baselinedec} taking advantage of the embedding $E_{nb}$ gathered from the neighbours. To do this, we use two different methods (Figure \ref{fig:model}).

First, we introduce the sequential aggregation where the neighbour information is mixed with the natural language information thanks to the cross-attention (as represented on the left of Figure \ref{fig:model}):

\begin{equation}
\label{eqn:sequentialaggregation}
\begin{split}
\text{DEC}(X, Y, \mathcal{N}) & = \text{Sublayer}^{\textsc{FFW}}(C_{nb}) \\
C_{nb} &=\text{Sublayer}^{\textsc{Cca}}(E_{nb}, C_{nl}) \\
E_{nb} &= \text{ENC\_NB}(\mathcal{N}, Y) \\
C_{nl} &=\text{Sublayer}^{\textsc{Ca}}(E_{nl}, C) \\
E_{nl} &= \text{ENC\_NL}(X) \\
C &= \text{Sublayer}^{\textsc{Sa}}(Y)
\end{split}
\end{equation}
The second solution computes the cross-attention between the neighbours and the natural language in parallel and then aggregate the information through a linear layer (as shown on the right of Figure \ref{fig:model}):
\begin{equation}
\label{eqn:parallelaggregation}
\begin{split}
\text{DEC}(X, Y, \mathcal{N}) & = \text{Sublayer}^{\textsc{FFW}}(C_{merge}) \\
C_{merge} &=\text{Linear}(C_{nb} + C_{nl}) \\
C_{nb} &=\text{Sublayer}^{\textsc{Cca}}(E_{nb}, C) \\
E_{nb} &= \text{ENC\_NB}(\mathcal{N}, Y) \\
C_{nl} &=\text{Sublayer}^{\textsc{Ca}}(E_{nl}, C) \\
E_{nl} &= \text{ENC\_NL}(X) \\
C &= \text{Sublayer}^{\textsc{Sa}}(Y)
\end{split}
\end{equation}
\footnotetext{Here, the neighbour's encoder is not constrained by the code being generated as in the original RETRO architecture and it is a classic transformer encoder. We drop it because it does not impact the results. See Appendix \ref{encneighboursappendix} for further details.}

The neighbour encoding provides a strong signal to the decoder, so we use equations \ref{eqn:sequentialaggregation} and \ref{eqn:parallelaggregation} every $p$ layers and otherwise we use the baseline equation \ref{eqn:baselinedec}.

\section{Dataset and preprocessing}
\label{setup}

In this Section, we describe the characteristics of the {\fontfamily{cmss}\selectfont CoNaLa} dataset on which we have tested our different architectures, the available code data to construct our database and the creation of the database.

\subsection{Dataset}

{\fontfamily{cmss}\selectfont CoNaLa} is a comprehensive corpus, comprising approximately 600,000 pairs of natural language expressions and their corresponding Python code fragments, sourced from StackOverflow. Among these examples, a subset of 2,879 pairs has undergone meticulous manual cleaning by professional developers, which significantly enhances their quality. This subset is further divided into a training set comprising 2,379 pairs and a fixed test set containing 500 pairs. 
All results reported in the article are based on these manually curated examples, unless stated otherwise. 
We created a fixed development set by extracting 200 examples from the total 2,379 examples within our training data.

CoNaLa was chosen due to its inclusion of a training set, distinguishing it from most recent datasets, like HumanEval \cite{humaneval}, which provide only evaluation sets. Our objective is to train models from scratch in a low-resource setting for comparative analysis with larger models, a goal not feasible with datasets lacking a training component.
\paragraph{Substitution} We  preprocess the CoNaLa dataset by normalising the names of variables and constants which are denoted by quotes in the natural language of the 2379 manually curated examples as done in \citep{tranX,BERTranX,docprompting}. This is done by substituting the actual names of the variables with a predefined set of normalized names that the statistical model can recognize. For example, all variables are renamed to \texttt{var\_0}, \texttt{var\_1}, etc. and all lists are renamed to \texttt{lst\_0}, \texttt{lst\_1}, etc. in both the natural language and code. 

\paragraph{Evaluation} To compare with previous work, we report the standard evaluation metric for {\fontfamily{cmss}\selectfont CoNaLa}. Hence, we report corpus-level BLEU and compare with other works on the fixed test set.

\subsection{Database creation}
\label{databasecreation}

The database's construction is the backbone of our model's framework. Here we delve into the specifics of the  dataset used for building our {\bf classic database} as described in \ref{classicdatabase}, and further explore the varied permutations surrounding this dataset.

Our intent is to harness the potential of the 600,000 code snippets extracted from CoNaLa. Given their intrinsic noise and sporadic alignment with natural language, these snippets raise a significant challenge when incorporated into model training. Nevertheless, the high volume of these snippets - corresponding to Python idiomatic tasks on StackOverflow - potentially holds a value for our model during its generation phase. We consistently draw from the totality of the 600,000 available codes with the 2,379 clean examples to construct our database. However, initial variations are introduced by modulating the length $m$ of the code, consequently leading to databases of different sizes (since each code is divided into $m$ chunks and each chunk corresponds to a single entry in the database). We vary the chunk length from 2, 4 and 8 (not counting the continuation which is of the same size of $m$) because in average code snippet from CoNaLa are of length $14.08$. Additionally, we introduce variations in the code snippets by integrating them 'as is' into the database or employing the substitution mechanism to standardize the codes by replacing the variable names (the heuristic for replace variable names and examples of substitution mechanism for mined examples are given in Appendix \ref{app:substitutionexamples}).

One limitation of the classical approach is the absence of constraints for the initially generated tokens. To leverage the statement and code snippet pairs, we opt to create a {\bf hybrid database} by integrating natural language embeddings as keys, along with the corresponding initial code segments as values. This integration guides the initial stages of our decoding process (Figure \ref{fig:CCA} in Appendix \ref{app:implementationdetails}). This variation aims to assist the model in generating the correct beginning of the code, which can be critical. Incorrect initial code sequences could lead to error propagation that becomes challenging to rectify in later generation steps, resulting in incorrect retrieval of neighbors as well.

We use codeBERT \footnote{https://github.com/microsoft/CodeBERT} to construct database embeddings and utilize Faiss \footnote{https://github.com/facebookresearch/faiss} to form the index.

\section{Experiments}
\label{experiments}

The experiments compare three strategies for code generation.   We start by describing our experimental protocol, highlighting the critical parameters utilized in our experiments\footnote{The code of our experiments is publicly accessible and can be found at {\tt https://github.com/NathanaelBeau/RET} \\{\tt ROseq2seq}.}. 
Then, we provide an analysis of the baseline transformer approach, thoroughly detailed in Section \ref{sec:baseline}.

To test the contribution of the database, we first evaluate an enhanced version of the baseline transformer with the {\bf classic database} (Section \ref{RETROcodebase}). This experiment is designed to investigate the effectiveness of augmenting the model with a broader context of code structure and familiar patterns.

Third, we investigate the {\bf hybrid database} approach, enhancing the database with natural language to constraint the decoding process at the beginning.

\subsection{Methodology}
Given the  amount of code at our disposal, we leverage \textsc{codeBERT} for natural language encoding, thereby ensuring its compatibility with our pre-existing seq2seq architecture. This approach is apt, considering \textsc{codeBERT}'s training not only involves code but also incorporates document strings corresponding to that code, thereby imbuing \textsc{codeBERT} with capabilities for understanding natural language.

For encoding and decoding tasks associated with the neighbours, we use 6-layer transformers equipped with 8 heads, maintaining hidden dimensions at a constant 256. We adhere to a fixed dropout of 0.4 across all cross-attention layers.


In all our experiments, we use two neighbours and use cross-attention every three layers as recommended by \citet{RETRO}. To optimize our model training, we precompute the neighbours during the database creation phase. Thus, our experimental strategy encompasses evaluating different chunk size configurations within the database, and also assessing the impact of variable name replacement. For the decoding process, a beam width of 15 was employed.



\subsection{Baseline}

Table \ref{table:baselinedev} summarizes the evaluation results of our two baseline configurations on our development set.

The first setup uses a system size of 168M parameters and yields a BLEU score of $35.19 \pm 0.63$ trained on the 2379 cleaned examples from {\fontfamily{cmss}\selectfont CoNaLa}. 

\begin{table}[htbp]
\centering
\scalebox{0.8}{
\begin{tabular}{|ll|c|}
 \hline
System & Size & BLEU  \\
\midrule
Baseline & 168M & $35.19 \pm 0.63$  \\ 
Baseline + 100k mined & 168M & $38.05 \pm 1.08$    \\
 \hline
\end{tabular}}
\caption{Baseline results on the development set.  The scores reported are the mean and standard deviation resulting from training with 5 different seeds.}
\label{table:baselinedev}
\end{table}

For the second configuration, we introduce an optimal addition of 100,000 mined examples, as recommended by \citet{docprompting}. The choice to exclude the full 600,000 mined examples from the training phase, even though they are used in our database, is based on the noted decline in performance. This decline is attributed to a misalignment between natural language and code, resulting from data noise, a conclusion supported by prior research \cite{tranX, BERTranX}. The integration of these mined examples enhances the model's performance, leading to a higher BLEU score of $38.05 \pm 1.08$.

The improvement observed in the 'Baseline + 100k mined' configuration highlights the effectiveness of augmenting the training set with mined examples. This observation supports the hypothesis that using mined examples can indeed serve as a significant strategy to improve the performance of code retrieval tasks. 

\subsection{RETROcode with classic database}
\label{resultclassic}

We now evaluate our models using the classic database to guide code generation. More specifically, we study the impact of different key variables, such as the substitution mechanism in the database, the chunk size $m$ (corresponding to the number of tokens in the database), and the method for aggregating the neighbours (either sequentially or in parallel), on the performance of our model.

\begin{table}[htbp]
\centering
\scalebox{0.75}{
\begin{tabular}{ |c|c|c|c| } 
\hline
Architecture & Substitution & chunk size $m$ & BLEU \\
\hline
\multirow{6}{*}{Parallel} & \multirow{3}{*}{False} & 2 & $32.98 \pm 0.93$  \\ \cline{3-4}
& & 4 & $28.53 \pm 1.05$ \\ \cline{3-4}
& & 8 & $31.56 \pm 0.72$ \\ \cline{2-4}
& \multirow{3}{*}{True} & 2 & $34.54 \pm 0.58$ \\ \cline{3-4}
& & 4 & $30.27 \pm 0.74$ \\ \cline{3-4}
& & 8 & $34.14 \pm 0.29$  \\ \cline{1-4}
\multirow{6}{*}{Sequential} & \multirow{3}{*}{False} & 2 & $34.35 \pm 0.36$  \\ \cline{3-4} 
& & 4 & $29.09 \pm 0.24$  \\ \cline{3-4} 
& & 8 & $31.71 \pm 0.49$  \\ \cline{2-4}
& \multirow{3}{*}{True} & 2 & $\boldsymbol{35.23 \pm 0.53}$ \\ \cline{3-4}
& & 4 & $31.59 \pm 0.67$  \\ \cline{3-4}
& & 8 & $34.60 \pm 0.65$ \\ \cline{3-4} 
\hline
\end{tabular}}
\caption{Comprehensive comparison of BLEU scores, each obtained from five different training sessions, on the development set by varying key parameters: system architecture (sequential or parallel), implementation of the substitution mechanism in the database, and the chunk size utilized in constructing the database. Each score represents a mean value along with the associated standard deviation.}
\label{table:RETROcodeclassic}
\end{table}

From Table \ref{table:RETROcodeclassic}, we observe that in all cases, the model performance is worse than that of our baseline, despite being trained on the same number of examples. A qualitative manual observation revealed that this disappointing behavior comes from generation errors at beginning of the sequence that are further propagated. The initial tokens of code are indeed generated without information from the neighbours (see Appendix \ref{app:inferenceclassicdb} for detailed output examples). The erroneous prefixes cause the query mechanism to retrieve similar beginning erroneous chunks, diverting our model from the correct path and consequently reducing the BLEU score significantly. From manual inspection again, we observe that the initial tokens of code generated are not fundamentally incorrect, but still different from the ground truth.

Before providing a solution to overcome this problem, let us first highlight the main trends 
for our different variables.

\paragraph{System Architecture} The results illustrate a significant variation between the parallel and sequential architectures. The sequential architecture appears to yield higher BLEU scores compared to the parallel one, particularly when the substitution mechanism is employed. It seems that cross-attention is a better way to merge information from natural language and neighbours rather than use a separate cross-attention treatment with a final linear layer.

\paragraph{Substitution} The implementation of a substitution mechanism consistently enhances the model's performance across both architectures and all chunk sizes. This increase in BLEU scores signifies that normalization of variable names through substitution can greatly help in retrieving appropriate neighbours and accurately predicting code. This is expected, given that one of the main difficulties in predicting code lies in predicting the variable name as described by \citet{BERTranX}. Furthermore, the retrieval of neighbouring codes is enhanced by substitution, which standardizes the code that has the same objective but uses different variable names.

\paragraph{Chunk Size} The impact of chunk size on model performance appears intricate, with no explicit pattern discernible from the Table. BLEU scores vary, not strictly correlating with the size of the chunks. For example, sometimes the smaller chunk size of 2 improves the results, likely by enabling the model to process more localized information from its neighbours. Conversely, larger chunk sizes, such as 8, also deliver good results as they let the model operate more independently during generation, with neighbouring data having less impact on the code's tail end. In the case of an intermediate chunk size of 4, however, the model seems to retrieve less relevant information, thus leading to confusion and potentially lower-quality code generation.

\subsection{RETROcode with hybrid database}
\label{hybriddatabase}

To avoid mismatches between the generated code and the reference code, we propose an initial stage of inference driven by a \textbf{hybrid database} build from  CoNaLa's clean and noisy pairs. By associating natural language embeddings (as key) with the beginnings of related codes (as value), we can query the database using the natural language input statement and retrieve corresponding code beginnings. This thereby guides the model from the generation's outset. The results for this method are detailed in Table \ref{table:RETROcodehybrid}.

\begin{table}[htbp]
\centering
\scalebox{0.75}{
\begin{tabular}{ |c|c|c|c| } 
\hline
Architecture & Substitution & chunk size $m$ & BLEU \\
\hline
\multirow{6}{*}{Parallel} & \multirow{3}{*}{False} & 2 & $35.87 \pm 0.71$  \\ \cline{3-4}
& & 4 & $32.22 \pm 0.38$ \\ \cline{3-4}
& & 8 & $36.81 \pm 1.07$ \\ \cline{2-4}
& \multirow{3}{*}{True} & 2 & $36.09 \pm 0.90$ \\ \cline{3-4}
& & 4 & $33.75 \pm 0.23$ \\ \cline{3-4}
& & 8 & $37.76 \pm 1.06$  \\ \cline{1-4}
\multirow{6}{*}{Sequential} & \multirow{3}{*}{False} & 2 & $39.10 \pm 0.79$  \\ \cline{3-4} 
& & 4 & $35.28 \pm 0.50$  \\ \cline{3-4} 
& & 8 & $43.03 \pm 1.18$  \\ \cline{2-4}
& \multirow{3}{*}{True} & 2 & $39.45 \pm 1.08$ \\ \cline{3-4}
& & 4 & $36.20 \pm 1.17$  \\ \cline{3-4}
& & 8 & $\boldsymbol{43.56 \pm 0.81}$ \\ \cline{3-4} 
\hline
\end{tabular}}
\caption{Exhaustive comparison of BLEU scores attained from five different training instances on the development set. Parameters echo those in Table \ref{table:RETROcodeclassic}, but with the initial database now augmented by the hybrid database; natural language embeddings (keys) are matched with the beginnings of corresponding codes (values). All scores represent means and corresponding standard deviations.}
\label{table:RETROcodehybrid}
\end{table}

The implementation of the hybrid database significantly enhances performance across all configurations. It notably achieves a BLEU score of $43.56$ with a chunk size of 8, exceeding the baseline + 100k by $5.5$ BLEU points. We observe empirically, that this method constraining generation from the very beginning often leads to codes closely resembling the ground truth, especially when m=8, allowing the model to frequently clone first neighbours that closely mirror the ground truth (see Appendix \ref{app:inferencehybriddb} for detailed output examples).

The observations for the different factors, as discussed in \ref{resultclassic}, remain consistent.

\subsection{Test set}

Finally we compare in table \ref{table:sotaconala} our best models against other state of the art systems on {\fontfamily{cmss}\selectfont CoNaLa} from \ref{hybriddatabase}. Additionally, to assess the robustness and general applicability of our model, we employ an alternative dataset, {\fontfamily{cmss}\selectfont CodeXGlue} \cite{codexglue}, with different properties. 

\paragraph{CoNaLa Test}

\begin{table}[htbp]
\scalebox{0.6}{
\begin{tabular}{ll|cc}
 \hline
System & Size & BLEU  & CodeBLEU \\
\midrule
ChatGPT-3.5-turbo \footnote{Evaluated on December 10, 2023. The date is specified to account for ongoing advancements in the ChatGPT model.} & ?B & {\bf 53.15}  & {\bf 60.50}  \\ 
Codex \citep{codex} & 12B & 43.16 & -   \\ 
CodeT5 + DocPrompting \citep{docprompting} & 220M & 36.22 & -\\
CodeT5 \citep{codet5} & 220M & 34.57 & -\\ 
kNN-BERTranX \cite{zhousyntax} & 240M & 37.29 & 39.04  \\
BERTranX \citep{BERTranX} & 130M & 34.20 & - \\ \midrule
RETROcode (parallel) + hybrid db  & 180M & 38.23  & 38.50  \\ 
RETROcode (sequential) + hybrid db & 176M & {\bf 43.09}  & {\bf 44.18 } \\ 
 \hline
\end{tabular}
}
\caption{Comparative analysis of system evaluated on the {\fontfamily{cmss}\selectfont CoNaLa}.}
\label{table:sotaconala}
\end{table}

We present result on Table \ref{table:sotaconala}. All systems use pre-training on external sources; for instance, we use \textsc{codeBERT} as the natural language encoder, whereas BERTranX utilizes BERT, and CodeT5, a seq2seq architecture, is pre-trained on the CodeSearchNet dataset \cite{codesearchnet}. ChatGPT and Codex, on the other hand, are pre-trained on a vast, undisclosed dataset. 
A unique strategy is seen in \citet{docprompting}'s approach, which enhances CodeT5's performance by incorporating additional information retrieved from a documentation database. BERTranX focuses on generating syntactically correct Python code through the construction of abstract syntax trees with a grammar-based decoder, while kNN-BERTranX enhances this with a grammar database. 
Our method, RETROcode, distinguishes itself in this competitive field by surpassing systems with similar scale and data sources by almost 5 BLEU points. It closely approaches the performance level of Codex, despite being significantly smaller in size — 66 times less than that of Codex — when tested on the {\fontfamily{cmss}\selectfont CoNaLa} dataset. However, it is important to note that our system still trails behind ChatGPT, which benefits from a considerably larger scale with possible data contamination and is more finely tuned for developer assistance.

\paragraph{CodeXGlue Test}

\begin{table}[htbp]
\scalebox{0.6}{
\begin{tabular}{ll|ccc}
 \hline
System & Size & BLEU  & CodeBLEU  \\
\midrule
ChatGPT-3.5-turbo\footnote{Evaluation made on December 15th 2023.}& ?B & {\bf 40.36}  & {\bf 54.47}   \\ 
Redcoder-Ext \cite{redcoder-ext} & 140M & 24.43 & 30.21 \\
GAP-Gen \citep{gap-gen} & 220M & 22.3 & 24.1 \\ \midrule
RETROcode (parallel) + hybrid db & 180M & 23.54  & 25.87 \\ 
RETROcode (sequential) + hybrid db & 176M & {\bf 27.41}  & {\bf 33.92} \\ 
 \hline
\end{tabular}
}
\caption{Comparative analysis of system evaluated on the {\fontfamily{cmss}\selectfont CodeXGlue}.}
\label{table:sotacodexglue}
\end{table}

The {\fontfamily{cmss}\selectfont CodeXGlue} dataset comprises 250,000 training examples and 15,000 test examples, each pairing a docstring with its corresponding Python function. This dataset poses a distinct challenge compared to development aid tasks, as it requires the generation of complete functions rather than mere one-liners. To adapt to this different coding requirement, we custom-build our database using the {\fontfamily{cmss}\selectfont CodeXGlue} training set, supplemented with examples from the Stack dataset \cite{thestack}. In this context, we maintain the use of two neighboring data points but increase the chunk size to $m=32$ to accommodate the complexity and length of the required code generation.
Redcoder-Ext also utilizes a code database, but its approach involves appending the retrieved code tokens directly to the input for processing through a pre-trained seq2seq model. Meanwhile, GAP-Gen advances Python code generation by emphasizing fine-tuning over pre-training and utilizes Syntax-Flow and Variable-Flow to guide its generation process.
In our assessments on the {\fontfamily{cmss}\selectfont CodeXGlue} dataset, our sequential RETROcode model demonstrates superior performance, surpassing Redcoder-Ext by nearly 3 BLEU points and 4 CodeBLEU points. This improvement is likely attributable to a more refined process of integrating neighboring data and managing the information flow.

For both datasets, we compute the $r(C)$ metric as utilized in RETRO which quantifies the overlap between test and database examples for both dataset. For {\fontfamily{cmss}\selectfont CoNaLa}, with $m=8$, we obtain a value of $r(C)=7.3\%$ while for {\fontfamily{cmss}\selectfont CodeXGlue} with $m=32$, we get $r(C)=10.2\%$.

\section{Conclusion}

In this paper, we introduced two novel seq2seq architectures to leverage natural language and a sizable code database for improved code generation. 
Our results reveal that the best way to integrate information from natural language and database neighbors is through direct cross-attention. We also identified the necessity to guide the initial stage of our generation process, achievable through a hybrid database that maximizes the benefits of the rich code resources and aligned pairs embedded in the dataset. 

\section*{Limitations}


A limitation arises from the model's scalability with database size, leading to longer computation times due to frequent database queries. For instance, our baseline model processes test examples in about 0.11 seconds each, whereas our improved model with a chunk size of 8 takes roughly 0.38 seconds per example. Despite this, the use of advanced libraries like SCANN and Faiss for k-nearest neighbors searches significantly enhances efficiency, enabling rapid querying of extensive databases (up to 2 trillion tokens) within 10 ms during evaluations.

The use of a static CodeBERT model presents another challenge, necessitating updates to accommodate new Python versions and software libraries. Nevertheless, CodeBERT's reliability and versatility ensure its effectiveness for our purposes, even without modifications specific to the task at hand.

Following the evaluation protocol used by {\fontfamily{cmss}\selectfont CoNaLa} and {\fontfamily{cmss}\selectfont CodeXGlue}, we use the BLEU and codeBLEU scores, but they do have inherent limitations. Firstly, the BLEU score does not vouch for the executability of the code - a single erroneous token can lead to a compilation error, despite high BLEU scores. Secondly, both scores do not accommodate for multiple viable codes capable of accomplishing the same task. In subsequent work, we plan to enrich our datasets and evaluation protocol with unit tests specifically testing the syntax and semantics of the generated code. The inclusion of such tests is expected to facilitate the formulation of more relevant metrics tailored for code generation evaluation.

Finally, the construction of the code database introduces a limitation. Our findings depend on the quality of the database, suggesting that future research could explore the impact of code quality. Additionally, care must be taken to exclude harmful or confidential code to mitigate security risks. Maintaining the database's content safety and integrity is crucial to prevent potential issues.

\bibliography{anthology,custom}
\bibliographystyle{acl_natbib}

\appendix

\section{Chunked cross-attention details}
\label{app:implementationdetails}

\begin{figure}[htbp]
\begin{center}
\scalebox{0.45}{
\includegraphics[width=\textwidth]{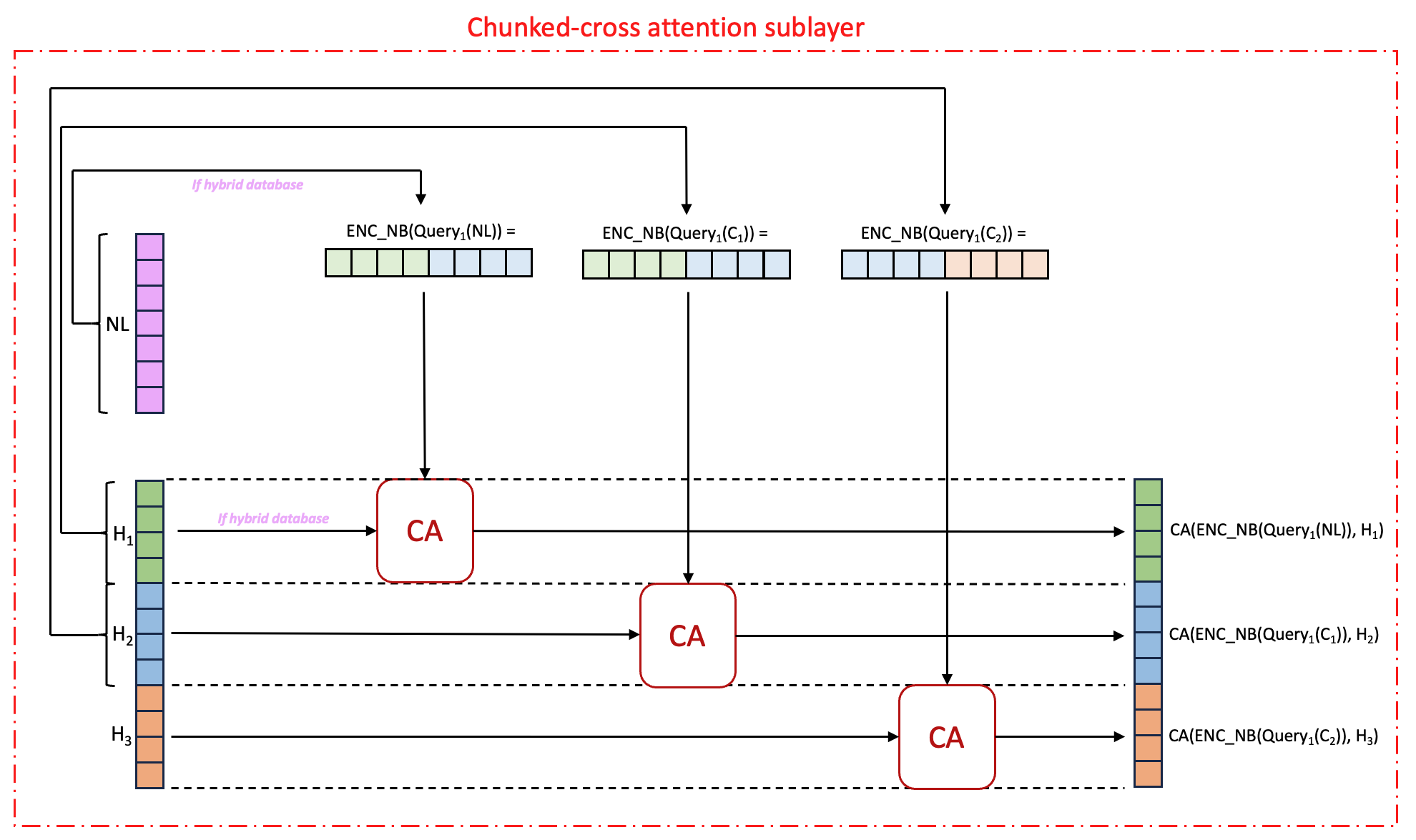}}
\end{center}
\caption{\label{fig:CCA}Illustration of chunk-cross attention mechanism with chunk length $m=4$. This illustration introduces a variation of the database, discussed in \ref{databasecreation}, featuring a hybrid database.}
\end{figure}

Our retrieval transformer model divides the output sequence into smaller segments, and uses information from previously processed segments to improve the accuracy of its predictions for the current segment. Specifically, the model retrieves text that is similar to the previous segment and uses this information to inform its predictions for the current segment. During training, it is important to not break autoregressivity of the model giving neighbours information too early to the decoder. To ensure that the model maintains autoregressivity during training, we use chunked-cross attention for neighours as in \citet{RETRO} where the input sequences are divided into smaller chunks, and the model performs cross-attention within each chunk.

Here's a breakdown of the process:
\begin{itemize}
    \setlength\itemsep{-0.1em}
    \item The input sequence is divided into smaller chunks. The sequence denoted as $Y$ is split into $l$ chunks, each of size $m$. This means that the hidden state $C$ is represented as a set of smaller chunks, denoted as $(C_u = (C_{u m+i})_{i \in \llbracket 1, m \rrbracket})$, where $u$ is an index that ranges from $\llbracket 1, l \rrbracket$.
    \item After the sequence is chunked, chunked cross-attention is computed between each chunk $C_u$ and its corresponding neighbour encodings $E_{nb, u}$. For each chunk $C_u$, for each token $i \in \llbracket 1, m \rrbracket$, we define:
\begin{equation*}
\textsc{Cca}(C, E_{nb}) = \textsc{Ca}(C_{u m+i}, E_{nb, u})
\end{equation*}
    \item There's a special case for the first $m$ tokens, which can't attend to any neighbour of a previous chunk. For these positions, cross-attention is defined as the identity. This means that for all tokens $j$ in the range from $1$ to $m$, the output of the chunked cross-attention operation is just the input itself, i.e., $\textsc{Cca}(C, E_{nb})_j = C_j$.
    \item    It's also important to note that this process is autoregressive, which means that the output of the chunked cross-attention operation at position $i$ depends on all the tokens from position $0$ to $i$ that have been input into the operation $\textsc{Cca}$.
\end{itemize}

This chunked cross-attention mechanism allows the model to handle sequences of data efficiently, by focusing on smaller chunks of the sequence at a time, while maintaining the ability to learn dependencies between different parts of the sequence through the cross-attention operation.

\section{Neighbours constrained encoder }
\label{encneighboursappendix}
The architecture proposed by \citet{RETRO} suggests that the generated code should constrain each layer of the neighbour encoder.

A straightforward strategy would be to use an encoder architecture akin to the natural language encoder, featuring two sub-layers; one for self-attention and one for feed-forward operations, as follows:
\begin{equation}
\label{eqn:neighbourencclassic}
\begin{split}
\text{ENC\_NB}(\mathcal{N}) & = \text{Sublayer}^{\textsc{FFW}}(H) \\
H &=\text{Sublayer}^{\textsc{Sa}}(\mathcal{N})
\end{split}
\end{equation}
We evaluate these two methods using our optimal model architecture - a sequential model with a hybrid database, as detailed in section \ref{hybriddatabase}. The database is preprocessed to normalize variable names, and it employs a chunk size of $m=8$. The results of this comparison are presented in Table \ref{table:constrainedencoder}.

\begin{table}[htbp]
\scalebox{0.7}{
\begin{tabular}{l|c|c}
\hline
System & Constrained Encoder & \textsf{CoNaLa} BLEU \\
\midrule
RETROcode (sequential) & False & $43.56 \pm 0.81$ \\
\hline
RETROcode (sequential) & True & $43.03 \pm 0.31$ \\
\hline
\end{tabular}}
\caption{Analysis of results from the development set, gathered from five distinct seeds, for our optimal model, both with and without the constrained neighbour encoder. Each BLEU score is expressed as an average value, accompanied by its corresponding standard deviation.}
\label{table:constrainedencoder}
\end{table}
 Interestingly, the BLEU scores show a marginal decrease when employing the constrained encoder approach. However, the standard deviation associated with the constrained method is notably lower, implying more consistent performance across different seeds. Hence, we decided to use a classical encoder for all experiments.

\mycomment{
\subsection{Neighbours encoder constrained thanks to natural language}

Neighbour encoding is a key point in our model, it is important to constrain it to get better results. Another possibility than the one proposed to constrain the encoding would be to use the natural language rather than the already generated code to encode the neighbours: 

\begin{equation}
\label{eqn:neighbourencca}
\begin{split}
\text{ENC\_NB}(\mathcal{N}, X) & = \text{Sublayer}^{\textsc{FFW}}(H) \\
H &=\text{Sublayer}^{\textsc{Ca}}(C, E) \\
C &= \text{Sublayer}^{\textsc{Sa}}(X)\\
E &= \text{Sublayer}^{\textsc{Sa}}(\mathcal{N}) 
\end{split}
\end{equation}
}

\section{Datasets mined examples}
\label{CoNaLaminedexamples}

We present in Table \ref{app:substitutionexamples} different examples from the CoNaLa mined examples used to construct our database.

\begin{table}[htbp]
\scalebox{0.75}{
\begin{tabular}{l|l}
 \hline
Intent & Snippet  \\
\midrule
\makecell{Convert binary string to list \\ of integers using Python} & \makecell{\texttt{[s[i:i + 3] for i in} \\ \texttt{range(0, len(s), 3)]}}   \\ 
\midrule
\makecell{How can I generate a \\ list of consecutive numbers?}  & \makecell{\texttt{list(range(9))}}    \\ 
\midrule
\makecell{Converting byte string \\ in unicode string}  & \makecell{\texttt{c.decode(} \\ \texttt{'unicode\_escape')}}    \\ 
\midrule
\makecell{Python: Get relative path \\ from comparing two \\ absolute paths}  & \makecell{\texttt{from os.path} \\\texttt{import relpath}}    \\ 
\midrule
\makecell{A python function that \\ accepts as an argument \\ either a scalar  or a \\ numpy array}  &  \makecell{\texttt{if isinstance(x, } \\ \texttt{np.ndarray):} \\ \texttt{return y}}   \\ 
\midrule
\makecell{Delimit a specific column \\ and add them as  \\ columns in CSV }  & \makecell{\texttt{df.join(c3)}}    \\ 
\midrule
\makecell{How can I find start and \\ end occurrence of \\ character in Python}  & \makecell{\texttt{df1 = df[df['test']} \\ \texttt{!=df['test'].} \\ \texttt{shift(+1)]}}    \\ 
\midrule
\makecell{Making multiple calls with \\ asyncio and adding result \\ to a dictionary}  & \makecell{\texttt{loop.run\_until\_complete} \\ \texttt{(asyncio.wait(tasks))}}    \\ 
\hline
\makecell{Python regex matching \\ in conditionals}  & \makecell{\texttt{match = patt.match(line)}}    \\ 
\hline
\makecell{How do I use \\ matplotlib autopct?}  & \makecell{\texttt{plt.show()}}    \\ 
\hline
\end{tabular}}
\caption{10 examples pick randomly from {\fontfamily{cmss}\selectfont CoNaLa} mined examples.}
\label{table:exampleconala}
\end{table}

\section{Database normalized examples}
\label{app:substitutionexamples}

As mentioned in the section  \ref{databasecreation}, we can detect and normalise the variable names of the codes to build the database. To detect variable names, we use the astor library \footnote{https://pypi.org/project/astor/} to transform each code snippet into an abstract syntax tree. Once completed, we browse the tree's leaves and retrieve the variable names, excluding those corresponding to library calls such as pandas or numpy. Examples of variable normalization are shown in Table \ref{table:substitutionmechanism}:

\begin{table}[htbp]
\centering
\scalebox{0.75}{
\begin{tabular}{c|c}
 \hline
Code & Normalized Code  \\
\midrule
\makecell{\texttt{results = [r for k} \\ \texttt{in keywords for r} \\ \texttt{in re.findall(} \\ \texttt{k, message.lower())]}} & \makecell{\texttt{var0 = [r for k} \\ \texttt{in var1 for r} \\ \texttt{in re.findall(} \\ \texttt{k, var2.lower())]}}   \\ 
\midrule
\makecell{\texttt{getattr(a, } \\ \texttt{'print\_test')()}} & \makecell{\texttt{getattr(var0,} \\ \texttt{'var1')()}}    \\ 
\midrule
\makecell{\texttt{json.dumps(geodata)}}& \makecell{\texttt{json.dumps(var0)}}    \\ 
\midrule
\makecell{\texttt{df.groupby([} \\ \texttt{df.index.date, 'action'])} \\ \texttt{.count()}} & \makecell{\texttt{var0.groupby([} \\ \texttt{str0.count()}}    \\ 
\hline
\makecell{\texttt{format(5e-10, 'f')}} & \makecell{\texttt{format(5e-10, 'var0')}}    \\ 
\hline
\end{tabular}}
\caption{5 examples pick randomly from {\fontfamily{cmss}\selectfont CoNaLa} mined examples before and after substitution mechanism}
\label{table:substitutionmechanism}
\end{table}

\section{Inference process for classic database}
\label{app:inferenceclassicdb}

\paragraph{Code generated at each time step $m$}

We showcase outputs at each time step where the model queries the database to provide a deeper understanding of the model's performance for each chunk size $m$. We exclusively display our top-performing models for each chunk size, corresponding to the sequential architecture coupled with a normalized database.

\paragraph{First example}  \mbox{}\\ Intent: {\it count the occurrences of item str0 in list var0} \\
Ground truth: \texttt{var0.count('str0')}

For $m=2$:
\begin{table}[htbp]
\centering
\scalebox{0.75}{
\begin{tabular}{|c|c|c|}
\hline
\textbf{$t$} & \textbf{Code Generated} & \textbf{Retrieved Neighbours} \\
\hline
\multirow{2}{*}{2} & \multirow{2}{*}{\texttt{<s>var0}} & \texttt{var0)[-1} \\
\cline{3-3}
 & & \texttt{2)[:, None} \\
\hline
\multirow{2}{*}{4} & \multirow{2}{*}{\texttt{<s>var0.count}} & \texttt{.count('/} \\
\cline{3-3}
 & & \texttt{.count('str0}\\
\hline
\multirow{2}{*}{6} & \multirow{2}{*}{\texttt{<s>var0.count('str0}} & \texttt{'str0')</s>} \\ 
\cline{3-3}
 & & \texttt{'str0')</s>}\\
 \hline
\multirow{2}{*}{8} & \multirow{2}{*}{\texttt{<s>var0.count('str0')</s>}} & - \\
\cline{3-3}
 & & -\\
 \hline
\end{tabular}}
\label{table:code_generation}
\end{table}

\vspace{\baselineskip}

For $m=4$:
\begin{table}[!htbp]
\centering
\scalebox{0.7}{
\begin{tabular}{|c|c|c|}
\hline
\textbf{$t$} & \textbf{Code Generated} & \textbf{Retrieved Neighbours} \\
\hline
\multirow{2}{*}{4} & \multirow{2}{*}{\texttt{<s>len(var0}} & \texttt{=len(var0 - 7)} \\
\cline{3-3}
 & & \texttt{<s>len(var0)</s><pad><pad>} \\
\hline
\multirow{2}{*}{7} & \multirow{2}{*}{\texttt{<s>len(var0)</s>}} & \texttt{-} \\
\cline{3-3}
 & & \texttt{-}\\
\hline
\end{tabular}}
\label{table:code_generation}
\end{table}

For $m=8$:

\begin{table}[htbp]
\centering
\scalebox{0.7}{
\begin{tabular}{|c|c|c|}
\hline
\textbf{$t$} & \textbf{Code Generated} & \textbf{Retrieved Neighbours} \\
\hline
\multirow{4}{*}{8} & \multirow{4}{*}{\multirow{2}{*}{\shortstack{\texttt{<s>count = sum(} \\\texttt{1 for item in}}}} & \texttt{<s>count = len([item for item} \\
 & & \texttt{in lst0 if item == str0]))} \\
 \cline{3-3}
& & \texttt{<s>count = lst0.count(} \\
& & \texttt{str0)</s><pad>} \\
\hline
\multirow{4}{*}{12} & \multirow{2}{*}{\multirow{2}{*}{\shortstack{\texttt{<s>count = sum(} \\\texttt{1 for item in} \\ \texttt{var0 if item} \\ \texttt{== str0)</s>}}}} & \texttt{-} \\
 & & \texttt{-} \\
 \cline{3-3}
& & \texttt{-} \\
& & \texttt{-} \\
\hline
\end{tabular}}
\label{table:code_generation}
\end{table}

\paragraph{Error Analysis of code generated}

For $m=2$, the model correctly generates the necessary code structure and accurately translates the given intent. The small chunk size allows the model to gradually build up the code, getting the initial variable at step 2, then adding the \texttt{.count} function at step 4, and finally adding the argument \texttt{'str0'} at step 6. It then correctly ends the sequence at step 8.

For $m=4$, the model fails to accurately translate the given intent. Model generates \texttt{len} keyword and retrieves out of context neighbours. 

For $m=8$, the model generates a more complex piece of code than the ground truth but it correctly implements the logic of the given intent. This shows that larger chunk sizes can lead to more complex but still accurate translations of the given intent. However, after step 8 the model fails to retrieve any useful neighboring chunks, which could again be due to the larger chunk size.

In summary, a smaller chunk size can be beneficial for the accuracy of simpler code sequences, while a larger chunk size can be better suited to generating more complex code. However, the latter can also lead to issues with retrieving useful neighboring chunks and correcting errors in the generated code.


\paragraph{Second example}  \mbox{}\\ Intent: {\it Joining data from dataframe var0 with data from dataframe var1 based on matching values of column str0 in both dataframes} \\
Ground truth: \texttt{var0.merge(var1, on='str0')}

For $m=2$:

\begin{table}[htbp]
\centering
\scalebox{0.7}{
\begin{tabular}{|c|c|c|}
\hline
\textbf{$t$} & \textbf{Code Generated} & \textbf{Retrieved Neighbours} \\
\hline
\multirow{2}{*}{2} & \multirow{2}{*}{\texttt{<s>pd}} & \texttt{<s>pd.con} \\
\cline{3-3}
 & & \texttt{<s>pd.con} \\
\hline
\multirow{2}{*}{4} & \multirow{2}{*}{\texttt{<s>pd.merge(}} & \texttt{ge(var0,} \\
\cline{3-3}
 & & \texttt{ge(var0,}\\
\hline
\multirow{2}{*}{6} & \multirow{2}{*}{\texttt{<s>pd.merge(var0, var1}} & \texttt{ var1)</s>} \\ 
\cline{3-3}
 & & \texttt{'str0')</s>}\\
 \hline
\multirow{2}{*}{8} & \multirow{2}{*}{\shortstack{\texttt{<s>pd.merge(var0, var1,} \\ \texttt{on}}}& \texttt{, on='k} \\ 
\cline{3-3}
 & & \texttt{, on=['}\\
  \hline
\multirow{2}{*}{10} & \multirow{2}{*}{\shortstack{\texttt{<s>pd.merge(var0, var1,} \\ \texttt{on=['}}} & \texttt{ =['lst0']} \\ 
\cline{3-3}
 & & \texttt{=['var0',}\\
  \hline
\multirow{2}{*}{12} & \multirow{2}{*}{\shortstack{\texttt{<s>pd.merge(var0, var1,} \\ \texttt{on=['str0',}}} & \texttt{str0', 'var1} \\ 
\cline{3-3}
 & & \texttt{=[var4', 'var2}\\ 
 \hline
 \multirow{2}{*}{14} & \multirow{2}{*}{\shortstack{\texttt{<s>pd.merge(var0, var1,} \\ \texttt{on=['str0', on='}}} & \texttt{  on='str0')} \\ 
\cline{3-3}
 & & \texttt{ on='str1')}\\
  \hline
 \multirow{2}{*}{16} & \multirow{2}{*}{\shortstack{\texttt{<s>pd.merge(var0, var1,} \\ \texttt{on=['str0', on='str0']}}} & \texttt{str0']</s>)} \\ 
\cline{3-3}
 & & \texttt{var4')]}\\
   \hline
 \multirow{2}{*}{18} & \multirow{2}{*}{\shortstack{\texttt{<s>pd.merge(var0, var1,} \\ \texttt{on=['str0', on='str0']</s>}}} & \texttt{-} \\ 
\cline{3-3}
 & & \texttt{-}\\
 \hline
\end{tabular}}
\label{table:code_generation}
\end{table}

For $m=4$:

\begin{table}[htbp]
\centering
\scalebox{0.7}{
\begin{tabular}{|c|c|c|}
\hline
\textbf{$t$} & \textbf{Code Generated} & \textbf{Retrieved Neighbours} \\
\hline
\multirow{2}{*}{4} & \multirow{2}{*}{\texttt{<s>s1 = }} & \texttt{<s>s1=pd.mer)} \\
\cline{3-3}
 & & \texttt{cols=str2)</s><pad><pad>} \\
\hline
\multirow{2}{*}{8} & \multirow{2}{*}{\texttt{<s>s1=pd.mer}} & \texttt{pd.merge(var0,} \\
\cline{3-3}
 & & \texttt{<s>pd.merge(var0,}\\
\hline
\multirow{2}{*}{12} & \multirow{1}{*}{\texttt{<s>s1=pd.merge(}} & \texttt{q(var0, var1, args} \\
\cline{3-3}
 & \texttt{var0,} & \texttt{array(str0, dtype=np}\\
\hline
\multirow{2}{*}{16} & \multirow{1}{*}{\texttt{<s>s1=pd.merge(}} & \texttt{var2, 'var3', var0} \\
\cline{3-3}
 & \texttt{var0, var1, '} & \texttt{var1, 'var2':var2}\\
\hline
\multirow{2}{*}{20} & \multirow{1}{*}{\texttt{<s>s1=pd.merge(}} & \texttt{-} \\
\cline{3-3}
 & \texttt{var0, var1, 'var1)</s>} & \texttt{-}\\
\hline
\end{tabular}}
\label{table:code_generation}
\end{table}

For $m=8$:

\begin{table}[htbp]
\centering
\scalebox{0.75}{
\begin{tabular}{|c|c|c|}
\hline
\textbf{$t$} & \textbf{Code Generated} & \textbf{Retrieved Neighbours} \\
\hline
\multirow{2}{*}{8} & \multirow{2}{*}{\texttt{<s>s1 = pd.mer}} & \texttt{<s>s1=pd.merge(var0,var1,how} \\
\cline{3-3}
 & & \texttt{<s>df1 = pd.read\_hdf('str0',} \\
\hline
\multirow{4}{*}{16} & \multirow{4}{*}{\multirow{2}{*}{\shortstack{\texttt{<s>s1=pd.merge(} \\\texttt{var0,var1,how}}}} & \texttt{ge(var0,var1,how=} \\
 & & \texttt{'inner',on['str0']} \\
 \cline{3-3}
& & \texttt{seq in zip(var0,} \\
& & \texttt{var0[1:])]</s><pad>} \\
\hline
\multirow{4}{*}{22} & \multirow{2}{*}{\multirow{2}{*}{\shortstack{\texttt{<s>s1=pd.merge(} \\\texttt{var0,var1,how=} \\ \texttt{'inner' ,on=} \\ \texttt{'str0')</s>}}}} & \texttt{-} \\
 & & \texttt{-} \\
 \cline{3-3}
& & \texttt{-} \\
& & \texttt{-} \\
\hline
\end{tabular}}
\label{table:code_generation}
\end{table}

\paragraph{Error Analysis of code generated}

This second example is longer and more complex than the first one. 

For $m=2$, we notice that the code begins to take shape from the second step with the initiation of the "pd" command, a familiar Pandas syntax. As the chunk size is quite small, the code is updated with high frequency, allowing the model to regularly revise its sequence based on new neighboring chunks. However, this approach has a drawback. The model has trouble creating longer, more complex code structures, possibly due to the small chunk size causing it to focus on smaller fragments of code rather than the overall structure.

For $m=4$, the model is making efficient use of the neighboring code chunks to generate a relatively more complex structure. However, it hallucinates at the end copying neighbours. The size of $m$ looks too small to handle complex structure and too big to locally update code. 

For $m=8$, the model correctly generates a more complex structure with a larger chunk size. It begins the "pd.merge" command at the 8th step and finishes the function with appropriate syntax at the 22nd step. However, the model copies the how argument, which was not requested, and as in the other cases, it does not start like the reference, which reduces the BLEU score.

In summary, there are trade-offs associated with the choice of chunk size. A smaller chunk size facilitates more frequent adjustments but may struggle with larger structures, while a larger chunk size may generate more complex code but lacks the granularity to refine it.

\section{Inference process for hybrid database}
\label{app:inferencehybriddb}

\paragraph{Code generated at each time step}
Here we give an example of the inference process with our best model only for $m=8$.

\paragraph{First example}  \mbox{}\\ Intent: {\it Convert a list of lists var0 to list of integers} \\
Ground truth: \texttt{var0=[int(''.join([str(y) for y in x])) for x in var0]}

\begin{table}[htbp]
\centering
\scalebox{0.7}{
\begin{tabular}{|c|c|c|}
\hline
\textbf{$t$} & \textbf{Code Generated} & \textbf{Retrieved Neighbours} \\
\hline
\multirow{4}{*}{0} & \multirow{4}{*}{\multirow{1}{*}{-}} & \texttt{<s>[int(''.join(str(d)} \\
 & & \texttt{ for d in x))} \\
 \cline{3-3}
& & \texttt{<s>[''.join(str(d) } \\
& & \texttt{for d in x) for x} \\
\hline
\multirow{4}{*}{8} & \multirow{2}{*}{\multirow{3}{*}{\shortstack{\texttt{<s>var0=[int(''.join(str(}}}} & \texttt{<s>[int(x) for x} \\
 & & \texttt{ in str(var0)]</s><pad><pad>]} \\
 \cline{3-3}
& & \texttt{<s>[''.join(str(d) } \\
& & \texttt{for d in x) for x} \\
\hline
\multirow{4}{*}{14} & \multirow{2}{*}{\multirow{3}{*}{\shortstack{\texttt{<s>var0=[int(''.join(str(} \\ \texttt{d for d in var0</s>}}}} & \texttt{-} \\
 & & \texttt{-} \\
 \cline{3-3}
& & \texttt{-} \\
& & \texttt{-} \\
\hline
\end{tabular}}
\label{table:code_generation}
\end{table}

\paragraph{Error Analysis of code generated}
The code here is well predicted with our hybrid database thanks to useful neighbours retrieved at $t=0$. It is interesting to note that even if the code is valid, the BLEU score is not equal to $100$ given the dummy variable \texttt{d} predicted by the model.

\end{document}